\documentclass[tablecaption=bottom,wcp]{jmlr} %

\usepackage{booktabs}
\usepackage[load-configurations=version-1]{siunitx} %

\theorembodyfont{\upshape}
\theoremheaderfont{\scshape}
\theorempostheader{:}
\theoremsep{\newline}

\jmlrproceedings{AABI 2020}{3rd Symposium on Advances in Approximate Bayesian Inference, 2020}

\title{Bayesian Evidential Deep Learning with PAC Regularization}

  \author{\Name{Manuel Hau\ss mann} \Email{manuel.haussmann@iwr.uni-heidelberg.de}\\
\addr HCI/IWR, Heidelberg University, Germany
\AND
\Name{Sebastian Gerwinn} \Email{sebastian.gerwinn@de.bosch.com}\\
\Name{Melih Kandemir} \Email{melih.kandemir@de.bosch.com}\\
\addr Bosch Center for Artificial Intelligence, Renningen, Germany}

\usepackage[utf8]{inputenc}
\usepackage[T1]{fontenc}  
\usepackage{xcolor}
\colorlet{darkblue}{blue!50!black}
\usepackage{url}           
\usepackage{booktabs}     
\usepackage{amsfonts}    
\usepackage{nicefrac}   
\usepackage{microtype} 
\usepackage{enumitem} 
\usepackage{graphicx}
\usepackage{amsmath,amssymb}
\usepackage{mathtools}
\usepackage{adjustbox}
\usepackage{dsfont}
\usepackage{wrapfig}
\newcommand{\mathbold}[1]{\ensuremath{\boldsymbol{\mathbf{#1}}}}
\usepackage{csquotes}

\DeclareRobustCommand{\KL}[2]{\ensuremath{\textrm{KL}\left(#1\;\|\;#2\right)}}

\DeclareRobustCommand{\Ep}[2]{\ensuremath{\mathds{E}_{#1}\left[#2\right]}}
\DeclareRobustCommand{\E}[1]{\ensuremath{\mathds{E}\left[#1\right]}}

\DeclareRobustCommand{\Var}[1]{\ensuremath{\mathrm{var}\left[#1\right]}}
\DeclareRobustCommand{\varp}[2]{\ensuremath{\mathrm{var}_{#1}\left[#2\right]}}
\DeclareRobustCommand{\var}[1]{\ensuremath{\mathrm{var}\left[#1\right]}}

\newcommand{\Dir}{\mathcal{D}\textit{ir}}

\newcommand{\InvGam}{\mathcal{I}\textit{nv}\mathcal{G}\textit{am}}
\newcommand{\Norm}{\mathcal{N}}

\newcommand{\Cat}{\mathcal{C}\textit{at}}

\newcommand{\diag}{\textrm{diag}}

\DeclareMathOperator*{\argmax}{arg\,max}

\newcommand{\mbf}{\mathbold{f}}

\newcommand{\mbh}{\mathbold{h}}

\newcommand{\mbm}{\mathbold{m}}

\newcommand{\mbs}{\mathbold{s}}

\newcommand{\mbw}{\mathbold{w}}
\newcommand{\mbx}{\mathbold{x}}
\newcommand{\mby}{\mathbold{y}}

\newcommand{\mbW}{\mathbold{W}}
\newcommand{\mbX}{\mathbold{X}}

\newcommand{\mbalpha}{\mathbold{\alpha}}

\newcommand{\mblambda}{\mathbold{\lambda}}
\newcommand{\mbmu}{\mathbold{\mu}}

\newcommand{\mbsigma}{\mathbold{\sigma}}

\newcommand{\mcD}{\mathcal{D}}

\newcommand{\mcH}{\mathcal{H}}

\newcommand{\mcO}{\mathcal{O}}

\allowdisplaybreaks

\usepackage{tikz}
\usepackage{pgfplots}
\usetikzlibrary{shapes.geometric, arrows, fit, automata, positioning, calc, arrows.meta}
\usetikzlibrary{backgrounds}
\tikzstyle{box} = [rectangle, minimum width=1.5cm, minimum height=1.5cm,text centered, draw=black, inner sep=7pt]
\tikzstyle{neuron} = [circle, minimum width=4mm, very thick, draw=blue!80!black]

\begin{document}

\maketitle

 \begin{abstract}
 We propose a novel method for closed-form predictive distribution modeling with neural nets. In quantifying prediction uncertainty, we build on Evidential Deep Learning, which has been impactful as being both simple to implement and giving closed-form access to predictive uncertainty. We employ it to model aleatoric uncertainty and extend it to account also for epistemic uncertainty by converting it to a Bayesian Neural Net.  While extending its uncertainty quantification capabilities, we maintain its analytically accessible predictive distribution model by performing progressive moment matching for the first time for approximate weight marginalization. The eventual model introduces a prohibitively large number of hyperparameters for stable training. We overcome this drawback by deriving a vacuous PAC bound that comprises the marginal likelihood of the predictor and a complexity penalty. We observe on regression, classification, and out-of-domain detection benchmarks that our method improves model fit and uncertainty quantification.
 \end{abstract}
 
\section{Introduction}
\label{sec:intro}

As the interest of the machine learning community in data-efficient and uncertainty-aware predictors increases, research on Bayesian Neural Networks (BNNs)~\citep{mackay1995probable, neal1995bayesian} gains prominence. Differently from deterministic nets, BNNs have stochastic weights. Thanks to their stacked structure, they propagate predictive uncertainty through the hidden layers and can characterize complex uncertainty structures. Exact inference of such a highly nonlinear system is analytically intractable and very hard to approximate with high precision. Consequently, most research on BNNs thus far focused on improving approximate inference techniques in terms of precision and computational cost~\citep{lobato2015probabilistic,kingma2015variational,louizos2017multiplicative}. These approaches take the posterior inference of global parameters as given and develop their approximation based on it.
A newly emerging alternative approach is direct predictive distribution modeling. It proposes devising a highly expressive predictive distribution with several free parameters. These parameters are then fit to data via maximum likelihood estimation. This way, observations are used to train directly the end product of interest: the predictive distribution, bypassing the need for an intractable posterior inference step. Some existing methods model the predictive distribution via a stochastic process parameterized as a neural net \citep{garnelo2018conditional}, while others introduce local priors on the likelihood activations, integrate them out and train the hyperparameters of the marginal \citep{sensoy2018evidential,  malinin2018prior}.

This work builds on Evidential Deep Learning (EDL) \citep{sensoy2018evidential}, due to its technical simplicity and observed effectiveness. EDL places a Dirichlet prior on the class assignment probabilities of a classifier and parameterizes the Dirichlet strengths of this prior with a neural net. While demonstrating substantial improvements in out-of-domain detection and adversarial robustness, it is not capable of decomposing epistemic and aleatoric uncertainties. 
We propose an efficient and effective method that extends EDL to BNNs, equipping it with more advanced uncertainty quantification and decomposition capabilities. We assume independent local weight random variables controlling the BNN for each input/target pair of data points, which share common hyperparameters. Marginalizing these data-point specific weights of our network we perform training via type 2 maximum likelihood (ML)~\citep{berger2013statistical} on the prior hyperparameters. This analytically intractable marginalization is approximated using the Central Limit Theorem (CLT). Differently from other weight marginalization approaches that assign global weight distributions on infinitely many neurons and recover Gaussian Processes (GP)~\citep{ neal1995bayesian,leep2018deep,alonso2019deep}, our formulation maintains finitely many hidden units per layer and assigns them individual weight distributions. Due to per data-point treatment of the weight marginalization the BNN scales linearly with the data size. 
The advantages of such a BNN with data-point specific marginalization comes at the expense of a major drawback. The number of hyperparameters in a weight-marginalized BNN grows proportionally to the number of synaptic connections. Maximizing the marginal likelihood w.r.t.\ such a large number of hyperparameters is prone to overfitting~\citep{bauer2016understanding}. Since the weight variables are marginalized out and their hyperparameters of the weight prior are set via optimization, the model can no longer incorporate regularizing knowledge other than the parametric form of the prior distribution (e.g.\ normal with mean and variance as free parameters). We address this drawback by deriving a provably vacuous Probably Approximately Correct (PAC)~\citep{macallester1999pac,macallester2003pac} bound that contains the marginal likelihood as its empirical risk term. Minimization of this PAC bound automatically balances the fit to the data and deviation from a prior regularizing hypothesis.

We compare our method on various standard regression, classification, and out-of-domain detection benchmarks  against state-of-the-art approximate posterior inference based BNN training approaches. We observe that our method provides competitive prediction accuracy and better uncertainty estimation scores than those baselines.

\section{Bayesian Evidential Deep Learning}

\paragraph{Evidential Deep Learning.} 
Classification with cross-entropy loss in deep learning can be interpreted as a categorical likelihood parameterized by a neural net. EDL generalizes this setup by parameterizing a prior by a neural net $f(\cdot;\mbw)$ with deterministic weights $\mbw$, instead of the likelihood. %
For classification, given a data set $\mcD=\{(\mbx_n,\mby_n)_{n=1}^N\}$ consisting of $N$ pairs of input $\mbx_n$ and target $\mby_n$, a natural choice for the prior $p(\mblambda_n|\mbw,\mbx_n)$ on the categorical likelihood is a Dirichlet distribution, and the final model becomes  
\begin{equation} 
    \mblambda_n|\mbx_n \sim  \Dir\big(\mblambda_n|\mbalpha_n\big),\qquad \mby_n|\mblambda_n \sim \Cat\big(\mby_n|\mblambda_n\big), \qquad\forall n, \label{eq:3factor_uncertainty}
\end{equation}
where $\mbalpha_n = f(\mbx_n; \mbw) + 1$. This way, the model explicitly accounts for the \emph{distributional uncertainty} which may arise due to a mismatch between the train and test data distributions. 
To train, the loss is the expected sum of squares between $\mby_n$ and $\mblambda_n$ with an additional regularizing Kullback Leibler divergence on the $\mblambda_n$.

\paragraph{Bayesian Local Neural Nets.} 
Parameterizing the likelihood by a BNN $f(\cdot;\mbw)$ with random variables $\mbw$ as the weights results in the following probabilistic model
\begin{equation*}
  \mbw_n \sim p_\phi(\mbw_n),\qquad \mby_n|\mbx_n,\mbw_n \sim p(\mby_n|\mbx_n, \mbw_n),\qquad \forall n
\end{equation*}
where $p(\mby_n|\mbx_n, \mbw_n)$ is some likelihood, %
and $p_\phi(\mbw_n)$ is a prior over local weights with shared hyperparameters $\phi$. Differently from a canonical BNN where all data points share the same global weight latent variable~\citep{blundell2015weight, gal2015bayesian, kingma2015variational, louizos2017multiplicative}, here the mapping between each input-output pair is determined by a separate random variable $\mbw_n$, giving a unique mapping constrained by sharing a common set of prior hyperparameters $\phi$, which is required for the \emph{type 2 ML} based objective we introduce below.
As this modification implies a collection of only local latent variables, we refer to the resulting model as a {\it Bayesian Local Neural Net (BLNN)}.
It consists of two sources of uncertainty. First, the \emph{model (epistemic) uncertainty} captured by the prior over the parameters, i.e.\ $p_\phi(\mbw_n)$, which accounts for the mismatch between the model and the true functional mapping from $\mbx_n$ to $\mby_n$. Second, the irreducible \emph{data (aleatoric) uncertainty} given by $p(\mby_n|\mbx_n,\mbw_n)$ stemming from irreducible measurement noise. %

\paragraph{BLNN Training with Type 2 ML and Prediction.} 
An alternative to full Bayesian model training with posterior inference is marginalizing out all latent variables and maximizing the marginal likelihood with respect to the hyperparameters, hence avoiding the posterior inference step on latent variables, referred to as \emph{type 2 ML} \citep{berger2013statistical}. %
The optimization objective then is: $\argmax_\phi \log p_\phi(\mby|\mbX)$.
Introducing an independent $\mbw_n$ for each pair $(\mbx_n,\mby_n)$ leads to a sum of $N$ independent marginal likelihoods, 
\begin{equation}
  \log p_\phi(\mby|\mbX) %
  =\sum_{n=1}^N \log \int p(\mby_n|\mbx_n,\mbw_n)p_\phi(\mbw_n)d \mbw_n = \sum_{n=1}^N \log p_\phi(\mby_n|\mbx_n) \label{eq:type2_objective}%
\end{equation}
which is amenable to using mini-batches for further scalability. %
The marginal of a training data point is identical to the posterior predictive for new test data $\mbx^*$. Hence, an analytic approximation developed for training is directly applicable to test time.

\paragraph{Analytic Marginalization of Local Weights with Moment Matching.} 
Marginalizing out the local weights $\mbw_n$ in \eqref{eq:type2_objective} is an intractable problem due to the highly nonlinear neural net appearing in the likelihood $p(\mby_n|\mbx_n,\mbw_n)$. However, we can marginalize the weights approximately by recursive moment matching resorting to the Central Limit Theorem (CLT). This technique has previously been used in BNNs for other purposes, such as expectation propagation~\citep{lobato2015probabilistic,ghosh2016assumed}, fast dropout~\citep{wang2013fast}, and variational inference~\citep{wu2018fixing}. We employ the same technique for marginalizing out the weights of the BLNN (see appendix).

\paragraph{Bayesian Evidential Deep Learning.} 
The original formulation for EDL builds on deterministic nets, hence assumes a point estimate on the weights (consciously ignoring model uncertainty) and a sum-of-squares loss term. We improve this framework by assigning a local prior on the EDL weights $p_\phi(\mbw_n),$\footnote{Throughout this work we assume ${\mbw_n \sim p_\phi(\mbw_n) = \Norm\big(\mbw_n|\mbmu,\diag(\mbsigma^2)\big)}$, i.e.\ $\phi = (\mbmu, \mbsigma^2)$.} and instead consider the optimization of the marginal likelihood, which amounts to employing a BLNN as a prior on the likelihood and marginalizing over all $\mbw_n$ as well as $\mblambda_n$.  We name the eventual model that combines BLNN with EDL {\it Bayesian Evidential Deep Learning (BEDL)}. By virtue of the localized weights, the marginal likelihood of BEDL factorizes across data points, bringing additive data point specific marginal log-likelihoods maintaining the central source of its scalability, formally
\begin{equation}
  \log p_\phi(\mby| \mbX) %
  = \sum_n \log \int p(\mby_n| \mblambda_n) p(\mblambda_n|\mbw_n, \mbx_n) p_\phi(\mbw_n) d \mblambda_n d \mbw_n = \sum_n \log p_\phi(\mby_n| \mbx_n).\label{eq:logmarginal_lik}
\end{equation}
The marginalization of $\mblambda_n$ on the last step can be performed analytically under conjugacy, efficiently approximated by Taylor expansion, or via Monte Carlo sampling, after moment matching to marginalize the weights~$\mbw_n$.
For the $C$-class classification task with one-hot encoded targets $\mby_n$ and Dirichlet distributed $\mblambda_n$, this gives us for the $n$-th term  in~\eqref{eq:logmarginal_lik}, 
\begin{equation}
  \log \int\int p(\mby_n|\mblambda_n)p(\mblambda_n|\mbalpha_n)d\mblambda_n\Norm(\mbf_n^L|\mbm_n,\mbs_n^2)d\mbf_n^L =\log\Ep{\Norm(\mbf_n^L|\mbm_n,\mbs_n^2)}{\prod_{c=1}^C\left(\frac{\alpha_{nc}}{\alpha_{n0}}\right)^{y_{nc}}},
  \label{eq:classification_marginalization}
\end{equation} 
where $\mbf^L_n$ is the last layer after marginalization of the others (see Appendix~D), and we use $\mbalpha_n =(\alpha_{n1},...,\alpha_{nC}) = \exp(\mbf_n^L)$, and $\alpha_{n0} = \sum_c \alpha_{nc}$.
Since the computational bottleneck on weight marginalizing is circumvented by the analytical CLT-based moment matching, the final expectation can be efficiently approximated by samples.

\section{A Vacuous PAC Bound to Regularize BEDL} 

Training the objective in \eqref{eq:logmarginal_lik} is effective for fitting a predictor on data. It also naturally provides a learned loss attenuation mechanism. However, it lacks a key advantage of the Bayesian modeling paradigm. As the hyperparameters of the weight priors are employed for model fitting, they no longer contribute to training as complexity penalizers. It is well-known from the GP literature that marginal likelihood-based training is prone to overfitting for models with a large number of hyperparameters~\citep{bauer2016understanding}.\footnote{A direct objection is that one could go one level higher in the hierarchy, introducing hyperpriors over the parameters $\phi$. We derive in the appendix how one could do this, but preliminary experiments have shown it to perform a lot worse than the PAC-based approach.} We address this shortcoming by complementing the marginal likelihood objective of \eqref{eq:logmarginal_lik} with a penalty term derived from learning-theoretic first principles. We tailor the eventual loss only for robust model training and keep it maximally generic across learning setups. This comes at the expense of arriving at a generalization bound that makes a theoretically trivial statement, yet brings significant improvements to training quality as illustrated in our experiments.

PAC bounds have been commonly used for likelihood-free and loss-driven learning settings. A rare exception by~\cite{germain2016pac} proves the theoretical equivalence of a particular sort of PAC bound to variational inference. Similarly, we keep the notion of a likelihood in our risk definition, but differently, we correspond our bound to the marginal likelihood. Given a predictor $h$ chosen from a hypothesis class $\mcH$ as a mapping from $\mbx$ to $\mby$, we define the true and the empirical risks as 
\begin{equation*}
  R(h) = - \Ep{\mbx,\mby \sim  \Delta}{p\big(\mby|h(\mbx)\big)}\qquad\text{and}\qquad R_\mcD(h)= -\frac{1}{N} \sum_{n=1}^N p\big(\mby_n|h(\mbx_n)\big),
\end{equation*}
for the data set $\mcD$ drawn from an arbitrary and unknown data distribution $\Delta$. The risks $R(h)$ and $R_{\mcD}(h)$ are bounded below by $-\max p(y|h(x))$ and above by zero. Although this setting relaxes the common assumption that bounds risk to the $[0,1]$ interval, it is substantially simpler than the one suggested in~\citep{germain2016pac}, which defines $R(h) = \Ep{x,y \sim  \Delta}{\log p(y|h(x))} \in (-\infty,+\infty)$. This unboundedness brings severe technical complications, which are no longer relevant for our approach. Denoting by $Q$ the distributions learnable over $\mcH$ and by $P$ some regularizing distribution on $\mcH$, according to Theorem~2.1\footnote{The theorem assumes risks to be defined within the $[0,1]$ interval in the original paper, but it is valid for any bounded risk. Our risk definitions can trivially be squashed into $[0,1]$ up to a constant.} in~\citep{germain2009pac} we have for any $\delta \in (0,1]$ and any convex function $d(\cdot,\cdot)$ 
\begin{align}
    \Pr\Bigg\{d&\Big(\Ep{h\sim Q}{R_\mcD(h)}, \Ep{h\sim Q}{R(h)}\Big)\leq \frac{\KL{Q}{P} + \log(B/\delta)}{N} \Bigg\}\geq 1 - \delta,
\end{align}
where $B= \Ep{\mcD\sim\Delta}{\Ep{h\sim P}{\exp\big(Nd(R_\mcD(h), R(h))\big)}}$. Using a quadratic distance measure $d(x,y) = (x - y)^2$ and suitably bounding $B$ by exploiting the boundedness of the likelihood and in turn the risk, we get as an upper bound on the expected true risk (see the appendix)
\begin{align}
    -\frac1N\sum_{n=1}^N \log p_\phi(\mby_n|\mbx_n) + \sqrt{\frac{\KL{Q}{P} - \log\delta}{N} + \frac{\log\max(B)}{N}},   
  \label{eq:pac_bound}
\end{align}
which is the objective we use to train BEDL that contains the marginal likelihood in the first term and a  regularizer in the second. The additional term resembles the KL term in the EDL loss, and gives a theoretically-grounded mechanism to incorporate regularization.  

\paragraph{Computation of the Bound for Classification.}
For a $C$-class classification we can compute the first term in \eqref{eq:pac_bound} as discussed above (\eqref{eq:classification_marginalization}). In the regularization term as we use (similar to EDL) $P = \Dir\big(\mblambda|(1,\ldots,1)\big)$, i.e.\ the assumption that each class is equally likely, as the regularizing distribution. Given $Q = \Dir(\mblambda|\mbalpha)$,  $\KL{Q}{P}$ is analytically tractable, as is the last term where we use the upper bound 
$\log \max(B)\le N$.

\paragraph{Looseness of the Bound.} It should be noted that while we use PAC theory to derive and motivate the final objective, it should no longer be used in its PAC interpretation, as the approximations result in a loose bound that is trivially fulfilled. Its justification lies mainly in its regularizing function.

\section{Experiments}

\begin{table*}%
  \centering
  \adjustbox{max width=\textwidth}{
    \begin{tabular}{lccccc}
      &  &   \multicolumn{1}{c}{MNIST} & \multicolumn{1}{c}{Fashion-MNIST} & \multicolumn{1}{c}{CIFAR~1-5} & \multicolumn{1}{c}{CIFAR~6-10} \\
      & & \multicolumn{1}{c}{(In Domain)} & \multicolumn{1}{c}{(Out-of-Domain)} & \multicolumn{1}{c}{(In Domain)} & \multicolumn{1}{c}{(Out-of-Domain)} \\
      \cmidrule(lr){3-4} \cmidrule(lr){5-6} & Reference & Test Error (\%) & ECDF-AUC &  Test Error(\%) & ECDF-AUC \\ \midrule
      MC Dropout  & \citep{gal2016dropout}    &          $1.12$ &          $0.429$   &          ${\bf 18.36}$ &          $0.946$  \\
      VarOut  & \citep{kingma2015variational}  &    $1.47$       &  $1.381$           &    $33.94$            &  $0.673$              \\
      DVI   & \citep{wu2018fixing}         &    $0.72$      & $1.318$      &    $23.32$      & $1.251$  \\ 
      EDL  & \citep{sensoy2018evidential}   &   $1.08$  &  $0.132$  & $20.34$  & $0.451$  \\
      \midrule
      BEDL & Ours     &       $0.81$    &  $1.512$              &   $24.38$        & $1.253$       \\
      BEDL+Reg & Ours    &           ${\bf 0.66}$                 &    ${\bf 0.055}$          &    $20.02$ & ${\bf 0.083}$  \\ \bottomrule
    \end{tabular}  
  }
  \caption{\textbf{Classification and OOD Detection.} Test error and area under the curve of the empirical CDF (ECDF-AUC) of the predictive entropies on two pairs of datasets.}\label{tab:class}\label{tab:ood}
\end{table*}

We evaluate  \emph{BEDL} and its PAC-regularized version \emph{BEDL+Reg} on several classification tasks. Complete details on the training procedure can be found in the appendix. Additionally we provide there a discussion of the computational cost as well as extensions to regression and a comparison to GP based approaches.

\paragraph{Classification and out-of-domain detection.}
We train LeNet-5 networks on the MNIST train split, evaluate their classification accuracy on the MNIST test split as the in-domain task, and measure their uncertainty on the Fashion-MNIST data set as the out-of-domain task, adhering to the protocol used in prior work \citep{louizos2017multiplicative, sensoy2018evidential}.\footnote{Due to the license status of the not-MNIST data conflicting with the affiliation of the authors, we have to change the setup of earlier work,  %
 using instead Fashion-MNIST as the closest substitute.} 
 We expect from a perfect model to predict true classes with high accuracy on the in-domain task and always predict a uniform probability mass on the out-of-domain task, i.e.\ the area under the curve of the empirical CDF (ECDF-AUC) of its predictive distribution entropy is zero. We perform the same experiment on CIFAR10 using the first five classes for the in-domain task and treating the rest as out-of-domain. We use $P:=\Dir\big(\mblambda|(1,\ldots,1)\big)$ as the regularization prior on the class assignment parameters, which has the uniform probability mass on its mean, encouraging an OOD signal in the absence of contrary evidence. In Table~\ref{tab:ood}, we compare BEDL+Reg against EDL~\citep{sensoy2018evidential}, also the non-Bayesian and heuristically trained counterpart of BEDL+Reg. We consider EDL also as a special case of Prior Networks \citep{malinin2018prior} that does not need to rely on OOD data during training time, commensurate for our training assumptions. We evaluate MC Dropout, VarOut, and DVI as baselines in this setup. BEDL+Reg improves the state of the art in all four metrics except the CIFAR10 in-domain task, where it ranks second after the prediction time weight sampling-based (hence less scalable) MC Dropout. Remarkably, BEDL+Reg detects the OOD samples with significantly better calibrated ECDF-AUC scores than EDL.

\section{Conclusion}\label{sec:conclusion}
We present a method for performing Bayesian inference within the framework of evidential deep learning. Employing type 2 maximum likelihood for inference and combining it with PAC-bounds for regularization, we achieve higher accuracy and better predictive uncertainty estimates while maintaining scalable inference. Exact inference in a fully Bayesian model such as a GP (c.f. Table~\ref{tab:reg}) or Hamiltonian Monte Carlo inference for BNNs~\citep{bui2016deep} are known to provide better error rates and test log-likelihood scores, yet their computational demand does not scale well to large networks and data-sets. Our method, on the other hand, shows strong indicators for improvement in uncertainty quantification and predictive performance when compared to other BNN approximate inference schemes with reasonable computational requirements. These benefits of the BEDL+Reg framework might especially be fruitful in setups such as model-based deep reinforcement learning, active learning, and data synthesis, where uncertainty quantification is a vital ingredient of the predictor. 

\bibliography{main}

\appendix

\section{Related Work}\label{sec:related-work}

\paragraph{CLT-based moment matching.}
The objective for variational inference on BNNs~\citep{kingma2015variational, wu2018fixing, gal2016dropout}, optimizing a global variational posterior $q(\mbw)$, consists of a computationally intractable $\Ep{q(\mbw)}{\log p(\mby|\mbx,\mbw)}$ that decomposes across data points. 
Fast dropout~\citep{wang2013fast} approximates these terms via local reparameterization with moment matching. The same local reparameterization has been later combined with a KL term to perform mean-field VI via MC sampling~\citep{kingma2015variational} or moment matching~\citep{wu2018fixing}. Our BLNN formulation is akin to an amortized VI approach learning a single global set of posterior parameters $\phi$ for the variational posterior approximation $q_{\phi}(\mbw)$. 
We use the same trick to marginalize the local weights, which keeps the machinery intact until the top-most step where the order of the $\log(\cdot)$ and $\E{\cdot}$ operations is swapped. This small change, however, has a large impact on the quality of uncertainty estimations.

\paragraph{Wide neural nets as GPs.} The equivalence of a GP to a weight-marginalized BNN with a single infinitely wide hidden layer has been discovered long ago~\citep{neal1995bayesian}. This result has later been generalized to multiple dense layers~\citep{leep2018deep,matthews2018gaussian}, as well as to convolutional layers~\citep{alonso2019deep}. The asymptotic treatment of the neuron count makes this approach exact at the expense of a lack of neuron-specific parameterization. The eventual GP has few hyperparameters to train, however, a prohibitively expensive covariance matrix to calculate. We employ the same training method on a middle ground where the hyperparameter count is double as many as a deterministic net and the cross-covariances across data points are not explicitly modeled.

\paragraph{Predictive density modeling.} 
Neural Processes~\citep{garnelo2018conditional, garnelo2018neural} follow a GP inspired approach of learning the predictive density using neural networks as part of the mapping from input to output space relying on an incorporation of the input data as context points for the predictive distribution of a test point. 
Earlier work on prior networks~\citep{malinin2018prior} 
parameterizes a prior to a classification-specific likelihood with deterministic neural nets, hence, discards model uncertainty. Additionally, they require samples from another domain to learn the distributional awareness. BEDL reformulates prior networks independently from the output structure, extends them to support also model uncertainty, and introduces a principled scheme for their training.

\section{Uncertainty Decomposition.}
Following~\citet{depeweg18a} one can use the law of total variance to decompose the predictive variance for one data point $(\mbx,\mby)$ as follows
\begin{equation*}
  \var{\mby|\mbx} = \varp{\mbw}{\E{\mby|\mbw, \mbx}} + \Ep{\mbw}{\var{\mby|\mbw,\mbx}},
\end{equation*}
where the first term, $\varp{\mbw}{\E{\mby|\mbw, \mbx}}$, focuses on the contribution to this predictive uncertainty by the variance over the network weights, i.e.\ the epistemic uncertainty, while the second, $\Ep{\mbw}{\var{\mby|\mbw,\mbx}}$, represents the remaining variance in the likelihood for the average weights.
After having marginalized over $\mblambda$, the mean and variance $\E{\mby|\mbw,\mbx}$, $\var{\mby|\mbw,\mbx}$, are analytically tractable with 
\begin{align*}
  \E{y_c|\mbw,\mbx} = \frac{\alpha_c}{\alpha_0}~~\text{and}~~\var{y_c|\mbw,\mbx} = \frac{\alpha_c}{\alpha_0}\Big(1 - \frac{\alpha_c}{\alpha_0}\Big).
\end{align*}
The EDL formulation allows for an analytic computation of the predictive variance, however as it considers only deterministic weights, it gets for learned parameters $\hat \mbw$
\begin{equation*}
  \var{\mby|\mbx} = \var{\mby|\hat \mbw, \mbx},
\end{equation*}
i.e.\ only a measure of the aleatoric uncertainty, lacking the epistemic. Our extension allows for the decomposition of the predictive uncertainty maintaining analytical tractability of the approximation to a great extent 
\begin{align*}
  \var{\mby|\mbx} &= \varp{\mbw}{\E{\mby|\mbw, \mbx}} \nonumber+ \Ep{\mbw}{\var{\mby|\mbw,\mbx}}\approx \varp{\mbf^L}{\E{\mby\big|\mbf^L, \mbx}} +  \Ep{\mbf^L}{\var{\mby\big|\mbf^L,\mbx}},
\end{align*}
where the final variance and expectation can be efficiently approximated with samples as discussed above.

\section{Extensions}\label{sec:app-hyper}
\subsection{BEDL with Hyperpriors}\label{sec:app-hyper}

Instead of relying on the PAC-bound based could try to incorporate further hyper-priors on $\phi$. This would hopefully reintroduce the missing regularization BEDL faces. The hierarchical model then has the following structure:
\begin{align*}
    \phi &\sim p(\phi),\\
    \mbw|\phi &\sim \prod_n p_\phi(\mbw_n),\\
    \mblambda|\mbw,\mbx &\sim \prod_np(\mblambda_n|\mbw_n,\mbx_n),\\
    \quad \mby|\mblambda &\sim \prod_n p(\mby_n|\mblambda_n).
\end{align*}
The marginal to be optimized over is then given as 
\begin{equation*}
    \log p(\mby,\phi|\mbX) = \sum_n \log \int p(\mby_n|\mblambda_n)p(\mblambda_n|\mbw_n,\mbx_n) p(\mbw_n|\phi)d\lambda_nd\mbw_n + \log p(\phi).
\end{equation*}
The first term is our regular marginal likelihood, while the second serves as as a regularizer as an optimization scheme aims to choose $\phi$ such that the marginal likelihood is high, but also that the prior density is large. The form of this hyperprior will vary depending on the problem at hand, but if we consider e.g.\ the $i$-th weight of the BNN $w_n^i$ to follow a normal distribution, we have
\begin{equation*}
    w_n^i|\phi_i \sim \Norm(w_n^i|\mu_i,\sigma^2_i),\quad\text{ where }\phi_i = (\mu_i,\sigma_i^2).
\end{equation*}
An obvious choice for the prior $p(\phi_i)$ is then given as
\begin{equation*}
    p(\phi_i) = p(\mu_i)p(\sigma_i^2) = \Norm(\mu|0,\alpha_0^{-1})\InvGam(\sigma^2|a_0,b_0).
\end{equation*}
The regression results summarized in Table~1 in the main paper however show that this approach tends to perform worse than both the PAC regularized BEDL as well as the unregularized BEDL.

\subsection{Generalization to Regression}\label{sec:application}

The EDL formulation was introduced by \cite{sensoy2018evidential} only for the case of classifications. However, the main motivation of the approach can also be extended to the case of regression.
We place a normal likelihood over the targets, treating the $\lambda_n$ as the mean and another normal as the distribution over $\lambda_n$ parameterizing both mean and variance with a BNN, giving  
\begin{align*}
  \mbw_n &\sim p_\phi(\mbw_n)\\
  \lambda_n | \mbw_n,\mbx_n &\sim \Norm\Big(\lambda_n \Big | f_1(\mbx_n;\mbw_n), \exp\big(f_2(\mbx_n;\mbw_n)\big)\Bigg)\\
  y_n | \lambda_n &\sim \Norm\left(y_n|\lambda_n, \beta^{-1}\right),
\end{align*}
with some fixed observation precision $\beta$. For the $n$-th sample in~\eqref{eq:logmarginal_lik} with $\mbf_n^L=(f_{n1}^L, f_{n2}^L)$, $\mbm_n=(m_{n1}, m_{n2})$, and $\mbs_n^2=(s_{n1}^2, s_{n2}^2)$ (the moment matching mean and variance), the log marginal likelihood is then given as 
\begin{equation*}
  \log p(y_n|\mbx_n) = \log \Norm\Big(y_n\Big|m_{n1}, \beta^{-1} + s_{n1}^2 + \exp(m_{n2}+s_{n2}^2/2)\Big).
\end{equation*}
The approximation is computed via a final moment matching step approximating the result of the inner integral $p(\lambda_n)=\int p(\lambda_n|\mbf_n^L)p(\mbf_n^L)d\mbf_n^L$ with a normal distribution while the ultimate equality follows directly from standard results on normal distributions. Contrary to the case of classification where the final step requires samples, the regression stays sampling-free.  We bound $\max{B}$ in~\eqref{eq:pac_bound} by exploiting that $\beta$ is fixed prior to training. Consequently, we get $\frac{\log\max(B)}{N}\le \frac{\beta}{2\pi}$ as a bound, which, as for the classification case, gives only a trivial performance guarantee (exceeding the maximum possible risk) but provides a justified training scheme.

\subsection{Further Extensions} 
Adaptations of CLT-based recursive moment matching to many other activation types and skip connections are feasible without further approximations. Max pooling can also be incorporated using approximations, but have also been shown to be replaceable altogether by strided convolutions without a performance loss~\citep{springenberg2015striving}. Deeper networks tend to require normalization procedures, which are not directly amenable to this moment matching. However tractable moments can also be computed for activation functions such as the ELU~\citep{clevert2015fast} as we show in the appendix alleviating this constraint.
The variance computations in \eqref{eq:varpreact} and the ReLU specific post-activation do not model any potential covariance structure between the pre-/post-activations units of a layer. While this is in principle feasible, e.g.\ along the lines of~\cite{wu2018fixing}, it leads to an explosion in the required computational cost and memory, hindering the applicability of the approach to deeper nets. Hence, we stick to a diagonal covariance structure throughout, as~\cite{wu2018fixing} have also shown only little test set performance benefit of modeling it.

\section{Further Details on the BLNN Derivations}\label{sec:app-blnn}

\paragraph{Derivation of the Moment Matching.}

For a single data point and the $l$-th hidden fully-connected layer\footnote{Convolutional layers follow analogously.}
consisting of $K$ units with an arbitrary activation function $a(\cdot)$, the post-activation layer output is given as $\mbh^l = a(\mbf^l)$, where ${\mbf^l = \mbW^l\mbh^{l-1}}$. The $j$-th pre-activation output $f_j^l$ is  a sum of $K$ terms
$f_j^l = \sum_k w_{jk}^lh_k^{l-1},$
which allows us to assume it to be normal distributed via the CLT due to the independence of the individual $w_{jk}^l$ and $h_k^{l-1}$ terms. The mean and the variance of this random variable can be computed as 
\begin{align}
  \E{f_j^l} &= \sum_{k=1}^K\E{w_{jk}^l}\E{h_k^{l-1}}, \\ 
  \var{f_j^l} &= \sum_{k=1}^K\E{(w_{jk}^l)^2}\var{h_k^{l-1}} + \var{w_{jk}^l}\Big(\E{h_k^{l-1}}\Big)^2, 
  \label{eq:varpreact}
\end{align}
where we drop any potential covariance structure between the outputs of a layer.
The mean and the variance of the weights are readily available via the distributions $p_\phi(\mbw_n)$. For common activations such as the ReLU, $a(h_k^{l-1}) = \max(0,h_k^{l-1})$, which we will rely on in this work, closed-form solutions to the first two moments of $h_k^{l-1}$ are tractable~\citep{frey1999variational} given the moments of the pre-activations of the previous layer $\mbf^{l-1}$.
This gives a recursive scheme terminating at the input layer
, where  $f^1_j = \sum_k w_{hj}^1x_k $. As $x_k$ is a constant, its first moment is itself and the second is zero. Consequently, 
\begin{equation*}
  \E{f_j^1} = \sum_{k=1}^K \mathds{E}[w_{jk}^1]x_k\quad\text{and}\quad \var{f_j^1}= \sum_{k=1}^K \mathrm{var}[w_{jk}^l] x_k^2,     
\end{equation*}
completing the full recipe of how all weights of a BNN can be recursively integrated out from bottom to top, subject to a tight approximation. 
Scenarios with stochastic input $\mbx \sim p(\mbx)$ typically entail controllable assumptions on $p(\mbx)$. The equations above remain intact after adding an expectation operator around $x_k$ and $x_k^2$, readily available for any explicitly defined $p(\mbx)$. Contrarily to the case in GPs, stochastic inputs can be trivially adapted into this framework, greatly simplifying the math for uncertainty-sensitive setups.
For a net with $L$ layers, the outcome is a distribution over the final latent hidden layer $\mbf^L_n \sim \Norm(\mbf^L_n|\mbm_n,\mbs_n^2)$, simplifying the highly nonlinear integrals in~\eqref{eq:type2_objective} to
\begin{equation*}
  \int p(\mby_n|\mbx_n,\mbw_n)p(\mbw_n) d\mbw_n \approx \int p(\mby_n|\mbf^L_n)\Norm(\mbf_n^L|\mbm_n,\mbs_n^2)d\mbf_n^L,
\end{equation*}
leaving us in a much simpler situation as we can choose a suitable distribution family for $\mby_n$.

\paragraph{First two Moments of the ReLU Activation.}
Mean and variance of a normally distributed variable transformed by the ReLU activation function are analytical tractable~\citep{frey1999variational}. Following the notation from the main paper, we have that for ${h_k^{l-1} = \max(0,f_k^l)}$ where $f_k^{l-1}\sim \Norm(f_k^{l-1}|\mu,\sigma^2)$ for some mean and variance they can be computed as 
\begin{align*}
  \E{h_k^{l-1}}&=\Ep{\Norm(f_k^{l-1}|\mu,\sigma^2)}{\max(0,f_k^{l-1})} = \mu\Phi\left(\frac\mu\sigma\right) + \sigma\phi\left(\frac\mu\sigma\right),\\
  \var{h_k^{l-1}}&=\var{\max(0,f_k^{l-1})} = (\mu^2 + \sigma^2)\Phi\left(\frac\mu\sigma\right)  + \mu\sigma\phi\left(\frac\mu\sigma\right) - \big(\E{h_k^{l-1}}\big)^2,\label{eq:varpostact}
\end{align*}
where $\Phi(\cdot)$ and $\phi(\cdot)$ are the cdf and pdf of the standard normal distribution respectively.

\paragraph{First two Moments of the ELU Activation.}

The following derivations are an adaptation of the ReLU results to ELU in order to scale to deeper networks. We are again interested in the first two moments for the ELU defined as 
\begin{equation}
  g(x) = \begin{cases}
    x, &x > 0\\
    \alpha(\exp(x) - 1) & x < 0\\
  \end{cases}.
\end{equation}
With $f(x) = \max(0,x)$, i.e.\ the ReLU activation, we have for the expectation of $g(\cdot)$, that 
\begin{equation*}
  \E{g(x)} = \int_{-\infty}^{0} \alpha(\exp(x) - 1) \Norm(x|\mu, \sigma^2) dx + \E{f(x)}.
\end{equation*}
The first term can be split into
\begin{equation*}
  \alpha \int_{-\infty}^{0} \exp(x) \Norm(x|\mu,\sigma^2)dx - \alpha\int_{-\infty}^{0} \Norm(x|\mu,\sigma^2).
\end{equation*}
We get for these to terms that the first is equal to
\begin{equation*}
  \alpha \int_{-\infty}^{0} \exp(x) \Norm(x|\mu,\sigma^2)dx = \alpha \exp(\mu + \sigma^2/2)\Phi\left (-\frac{\mu + \sigma^2}{\sigma}\right )
\end{equation*}
and the second gives
\begin{align*}
  \alpha\int_{-\infty}^{0} \Norm(x|\mu,\sigma^2) = \alpha \Phi\left (-\frac\mu\sigma\right ) =  \alpha \left (1 - \Phi\left (\frac\mu\sigma\right )\right ). 
\end{align*}
Combining all of this we end up with
\begin{equation*}
  \E{g(x)} = \alpha \left(\exp(\mu + \sigma^2/2)\Phi\left (-\frac{\mu + \sigma^2}{\sigma}\right ) -   \Phi\left (-\frac\mu\sigma\right )      \right) + \E{f(x)}.
\end{equation*}
For the variance we have the general form 
\begin{equation*}
  \Var{g(x)} = \E{g(x)^2} - \E{g(x)}^2
\end{equation*}
in which only the second moment is missing. 
We have that 
\begin{equation*}
  \E{g(x)^2} = \int_{-\infty}^{0} \alpha^2(\exp(x) - 1)^2\Norm(x|\mu,\sigma^2)dx + \E{f(x)^2}.
\end{equation*}
The first term expands into the following monstrosity
\begin{align*}
  &\alpha^2 \int_{-\infty}^{0}(\exp(2x) - 2\exp(x) + 1) \Norm(x|\mu,\sigma^2)dx\\
  =\quad&\alpha^2\int_{-\infty}^{-\frac\mu\sigma}\big(\exp(2\mu + 2\sigma y) - 2 \exp(\mu+ \sigma y) + 1\big)\phi(y)dy\\
  =\quad&\alpha^2\left(\exp(2\mu + 2\sigma^2)\Phi\left(-\frac{\mu + 2\sigma^2}{\sigma} \right) \right.\\
  &\quad\left.- 2\exp(\mu + \sigma^2/2)\Phi\left(-\frac{\mu+\sigma^2}\sigma\right) + \Phi\left(-\frac\mu\sigma\right)\right).
\end{align*}
These two moments finally can be used as replacements for the ReLU moments in the main paper for deeper networks.

\paragraph{Derivation of the Marginalization for Regression.}\label{sec:app-margreg}
We approximate the marginal distribution 
\begin{align*}
  p(\lambda_n) = \int\Norm\big(\lambda_n|f_{n1}^L, \exp(f_{n2}^L)\big)\Norm(\mbf_n^L|\mbm_n,\mbs_n^2)d\mbf_n^L
\end{align*}
with a normal distribution by a further moment matching step. Dropping the indices $n$ and $L$ for notational simplicity, the mean of the right hand side is given as 
\begin{align*}
  \Ep{p(\lambda)}{\lambda} &= \int \lambda p(\lambda)d\lambda= \int \lambda \Norm(\lambda|f_1,\exp(f_2))\Norm(\mbf|\mbm,\mbs^2)d\mbf d\lambda\\
  &=\int f_1 \Norm(\mbf|\mbm, \mbs^2)d\mbf = m_1.
\end{align*}
For the variance term we rely on the law of total variance and have 
\begin{align*}
  \varp{p(\lambda)}{\lambda} &= \Ep{p(\mbf)}{\varp{p(\lambda|\mbf)}{\lambda}} + \varp{p(\mbf)}{\Ep{p(\lambda|\mbf)}{\lambda}}\\
  &=\Ep{p(\mbf)}{\exp(f_2)} + \varp{p(\mbf)}{f_1}\\
  &= \int \exp(f_2) \Norm(f_2|m_2,\sigma_2^2)df_2 + s_1^2\\
  &=\exp(m_2 + s_2^2/2) + s_1^2,
\end{align*}
where the last integral is given as the mean of a log-normal random variable. Altogether we end up with the desired
\begin{equation*}
  p(\lambda_n) \approx \Norm\Big(\lambda_n\big|m_{n1}, s_{n1}^2 + \exp(m_{n2}+s_{n2}^2/2)\Big).
\end{equation*}

This then allows us to compute the log marginal likelihood
\begin{align*}
  \log p(y_n|\mbx_n)&=\log \int  \Norm(y_n|\lambda_n,\beta^{-1}) \left(\int\Norm\big(\lambda_n|f_{n1}^L, \exp(f_{n2}^L)\big)\Norm(\mbf_n^L|\mbm_n,\mbs_n^2)d\mbf_n^L\right)d\lambda_n\nonumber\\
  &\approx \log \int \Norm(y_n|\lambda_n,\beta^{-1}) \Norm\Big(\lambda_n\big|m_{n1}, s_{n1}^2 + \exp(m_{n2}+s_{n2}^2/2)\Big) d\lambda_n\nonumber\\ %
  &= \log \Norm\Big(y_n\big|m_{n1}, \beta^{-1} + s_{n1}^2 + \exp(m_{n2}+s_{n2}^2/2)\Big).\nonumber
\end{align*}

\section{Derivation of the PAC-Bound}\label{sec:app-pac}

This section gives a more detailed derivation of the individual results stated in the main paper. As stated there, given a predictor $h$ chosen from a hypothesis class $\mcH$ as a mapping from $\mbx$ to $y$, we define the true and the empirical risks as 
\begin{align}
    R(h) &= - \Ep{\mbx,\mby \sim  \Delta}{p\big(\mby|h(\mbx)\big)},\\
    R_\mcD(h) &= -\frac{1}{N} \sum_{n=1}^N p\big(\mby_n|h(\mbx_n)\big)
\end{align}
for the data set $\mcD$ drawn from an arbitrary and unknown data distribution $\Delta$. $R(h)$ and $R_{\mcD}(h)$ are bounded below by $-\max p(y|h(x))$ and above by zero. 

Theorem~2.1 in~\citep{germain2009pac} gives us that for any $\delta \in (0,1]$ and any convex function $d(\cdot,\cdot)$ 
\begin{align}
    \mathrm{Pr} &\Bigg\{d\Big(\Ep{h\sim Q}{R_\mcD(h)}, \Ep{h\sim Q}{R(h)}\Big)\leq\frac{\KL{Q}{P} + \log(B/\delta)}{N}) \Bigg\}\geq 1 - \delta,
\end{align}
where $B= \Ep{\mcD\sim\Delta}{\Ep{h\sim P}{\exp\big(Nd(R_\mcD(h), R(h))\big)}}$. The PAC framework necessitates a convex and non-negative distance measure for risk evaluations. Common practice is to rescale the risk into the unit interval, define the KL divergence as the distance measure, and upper bound its intractable inverse~\citep{germain2016pac} using Pinsker's inequality~\citep{catoni2007pac, dziugaite2017computing}. We follow an alternative path. As our risk is bounded but not restricted to the unit interval, we choose our distance measure as $d(r,r') = (r-r')^2$ and avoid the Pinsker's inequality step. 

Adapting the standard KL inversion trick~\citep{seeger2002pac} to the Euclidean distance, we can simply define $d^{-1}(x,\varepsilon) = \max \{ x' : (x-x')^2 = \varepsilon \}= x + \sqrt{\varepsilon}$
for some $\varepsilon \geq 0$. We apply this function to both sides of the inequality and get
\begin{align*}
  &\quad d^{-1} \Big(\Ep{h \sim Q}{R_\mcD(h)},d\big(\Ep{h \sim Q}{R_\mcD(h)},\Ep{h \sim Q}{R(h)}\big)\Big) \\
  &\leq d^{-1} \Big(\Ep{h \sim Q}{R_\mcD(h)},\big(KL (Q || P)+ \log (B/\delta)\big)/N\Big),
\end{align*}
where $d(\cdot,\cdot) \geq 0$ and $KL (Q || P) \geq 0$ by definition and because $\delta \in [0,1]$ and $e^{d(\cdot,\cdot)} \geq 0$ we have $\log (B/\delta) = -\log\delta + \log B   \geq 0$ . Since 
\begin{equation*}
    \Ep{h \sim Q}{R(h)}  \leq \Ep{h \sim Q}{R(h)}+ d\Big(\Ep{h \sim Q}{R_\mcD(h)},\Ep{h \sim Q}{R(h)}\Big)
\end{equation*} 
directly follows from $ d^{-1}(\cdot,\cdot)$, we bound the true risk as
\begin{equation*}
   \mathrm{Pr}  \Bigg \{  \Ep{h \sim Q}{R(h)} \leq\Ep{h \sim Q}{R_\mcD(h)} + \sqrt{ \frac{KL(Q||P) + \log (B/\delta)}{N}  } \Bigg \} \geq 1-\delta.
\end{equation*}
This outcome has a similar structure to application of Pinsker's inequality to a setup with risk defined on the unit interval, but without such a restriction. Hence, the implied upper bound is no longer trivial.
To arrive at the final bound we have to further approximate each of the two terms of the bound. 

For the first term, we have that
\begin{align*}
    \Ep{h\sim Q}{R_\mcD(h)} &= -\frac1N\sum_{n=1}^N\Ep{h\sim Q}{p\big(\mby_n|h(\mbx_n)\big)}\\
    &\leq -\frac1N\sum_{n=1}^N\log\Ep{h\sim Q}{p\big(\mby_n|h(\mbx_n)\big)} \\
    &= -\frac1N\sum_{n=1}^N\log p(\mby_n|\mbx_n),
\end{align*}
where the inequality uses that $-\log(u) > -u~\forall u \in (0,\infty)$ and the equality follows by the marginalisation techniques discussed in the main paper.

To get a tractable second term we process $B$ further. Exploiting the fact that 
\begin{align}
\Ep{x \sim p(x)}{\max f(x)} \leq \max f(x)
\end{align}
for any $p(\cdot)$ and $f(\cdot)$, we can drop the the expectation term and get
\begin{align}
    B&=\Ep{\mcD\sim\Delta}{\Ep{h \sim P} {e^{N(R_\mcD(h)-R(h))^2}}}  \nonumber \\
    &\leq\max e^{N(R_\mcD(h) - R(h))^2}.
\end{align}
For a multiclass classification, the likelihood is bounded into the interval $[0,1]$ such that with $\max (R_\mcD(h)-R(h))^2 = (0-1)^2 = 1$ we have that 
\begin{align}
    B \leq e^N \quad \Rightarrow\quad \log B \leq N.
\end{align}
For regression, with the likelihood $\Norm(y|h(x),\beta^{-1})$, we have that $R_\mcD(h)$ and $R(h)$ are bounded from above by $0$ and from below by $-\Norm\big(y=\mu|\mu,\beta^{-1}\big)$, i.e.\ by the density at the mode of a normal distribution with precision $\beta^{-1}$. Hence,
\begin{align}
    B \leq e^{N\big(0 - \Norm(\mu,\beta^{-1})\big)^2}\quad \Rightarrow\quad \log B \leq N \frac\beta{2\pi}. 
\end{align}
Combining these relaxations we get the objectives described in the main paper.

\section{Experimental Details and Further Experiments}\label{sec:app-expdetail}
This section contains further experiments as well as details on the hyperparameters of the experiments performed in the main paper. 

\subsection{Regression}

\begin{table*}
  \centering
  \adjustbox{max width=\textwidth}{
    \begin{tabular}{lrrrrrrrr}
      \toprule
      & \multicolumn{1}{c}{ \texttt{boston}} & \multicolumn{1}{c}{\texttt{concrete}} & \multicolumn{1}{c}{\texttt{energy}} & \multicolumn{1}{c}{\texttt{kin8nm}} & \multicolumn{1}{c}{\texttt{naval}} & \multicolumn{1}{c}{\texttt{power}} & \multicolumn{1}{c}{\texttt{protein}} & \multicolumn{1}{c}{\texttt{wine}}  \\
      \multicolumn{1}{c}{$N/d$}  &         \multicolumn{1}{c}{$506/13$} &          \multicolumn{1}{c}{$1030/8$} &         \multicolumn{1}{c}{$768/8$} &        \multicolumn{1}{c}{$8192/8$} &     \multicolumn{1}{c}{$11934/16$} &       \multicolumn{1}{c}{$9568/4$} &        \multicolumn{1}{c}{$45730/9$} &     \multicolumn{1}{c}{$1599/11$} \\ \midrule
      Sparse GP& $-2.22 \pm 0.07$ & $-2.85 \pm 0.02$ & $-1.29 \pm 0.01$ & $1.31 \pm 0.01$ & $4.86 \pm 0.04$ & $-2.66 \pm 0.01$ & $-2.95 \pm 0.05$ & $-0.67 \pm 0.01$   \\ \midrule
      MC Dropout & $-2.46\pm0.25$ &          $-3.04\pm0.09$ &          $-1.99\pm0.09$ &          $0.95\pm0.03$ &          $3.80\pm0.05$ &          $-2.89\pm0.01$ &          $-2.80\pm0.05$ &          $-0.93\pm0.06$ \\
      VarOut  &          $-2.63\pm0.02$ &          $-3.15\pm0.02$ &          $-3.29\pm0.00$ &          $1.09\pm0.01$ &          $5.50\pm0.03$ &          $-2.82\pm0.01$ &          $-2.90\pm0.01$ &         $ {\bf -0.88\pm0.02}$  \\
      PBP       &        $-2.57 \pm 0.09$ &          $-3.16\pm0.02$ &         $-2.04\pm 0.02$ &         $0.90\pm 0.01$ &          $3.73\pm0.01$ &         $-2.84\pm 0.01$ &         $-2.97\pm 0.00$ &          $-0.97\pm0.01$ \\      
      DVI        &          $\mathbf{-2.41\pm0.02}$ &          $-3.06\pm0.01$ &          $-1.01\pm0.06$ &          $1.13\pm0.00$ &          $\mathbf{6.29\pm0.04}$ &          $-2.80\pm0.00$ &          $-2.84\pm0.01$ &          $-0.90\pm0.01$  \\
      \midrule 
      BEDL-Hyper (Ours)  &         $-2.57\pm 0.04$ &         $-3.30 \pm0.01$ &          $-2.59\pm0.02$ &          $0.44\pm0.00$ & $3.69\pm0.00$ &         $-2.98\pm 0.01$ &       $ -3.00 \pm 0.00$ &        $-1.00 \pm 0.01$ \\ 
      BEDL (Ours)     &    $-2.45 \pm 0.08$     & $-3.09 \pm 0.06$ &   $-0.87  \pm 0.10$    &  $1.12\pm0.01$ & $5.76 \pm 0.07$ & $-2.80 \pm 0.01$  & $-2.82 \pm 0.01$   & $-0.93 \pm 0.01$    \\
      BEDL+Reg (Ours)    &  $-2.43 \pm 0.06$       & ${\bf  -3.02 \pm 0.02}$  &  ${\bf -0.73  \pm 0.04}$     & ${\bf 1.15 \pm 0.01}$   & $5.60 \pm 0.11$  &  ${\bf -2.79 \pm 0.01}$ &  ${\bf -2.77 \pm 0.01}$  &  $-0.90 \pm 0.01$ \\
      \bottomrule
  \end{tabular}}
  \caption{\textbf{Regression.} Average test log-likelihood $\pm$ standard error over 20 random train/test splits. N/d give the number of data points in the complete data set and the number of input feature. The sparse GP results are cited from~\citep{bui2016deep} and VarOut relies on our own implementation.}\label{tab:reg}
\end{table*}

We evaluate the regression performance of BEDL+Reg and the baselines on eight standard UCI benchmark data sets. Adopting the experiment protocol introduced in~\citep{lobato2015probabilistic}, we use 20 random train-test set splits comprising $90\%$ and $10\%$ of the samples, respectively. The nets consist of a single hidden layer with 50 units and ReLU nonlinearities.\footnote{Except for the larger \texttt{protein}, which gets 100 hidden units.} The hypothesis class in this task is over the regularization parameters  $P:=p(\mblambda) = \prod_n\Norm(\lambda_n|0,\alpha^{-1})$, with precision $\alpha$, while $Q$ is given as $Q:=p(\mblambda)=\prod_n\int p(\lambda_n|\mbf_n)p(\mbf_n)d\mbf_n$. 
We compare BEDL+Reg against the state of the art in BNN inference methods that do not require sampling across  neural net weights, \emph{Probabilistic Back-Propagation} (PBP)~\citep{lobato2015probabilistic} and \emph{Deterministic Variational Inference} (DVI)~\citep{wu2018fixing}, which use the CLT-based moment matching for expectation propagation and VI, respectively.
For completeness, we also compare against the two most common sampling-based alternatives, \emph{Variational Dropout}~(VarOut)~\citep{kingma2015variational, molchanov2017variational} and \emph{MC~Dropout}~\citep{gal2016dropout}.
In the results summarized in Table~\ref{tab:reg}, BEDL+Reg outperforms all baselines in the majority of the data sets and is competitive in the others. 

The PAC regularization improves over BEDL in all data sets except one. We also report results for a sparse GP with 50 inducing points, which approximates a BNN of one infinitely wide hidden layer \citep{neal1995bayesian}. As expected, the GP sets a theoretical upper bound on BEDL+Reg as well as the baselines for one hidden layer architectures. Lastly, we compare our tediously derived PAC regularizer to straightforward Maximum-A-Posteriori estimation on the BEDL hyperpriors~(BEDL-Hyper) (see the appendix for details), which deteriorates performance on all UCI data sets.

\paragraph{Experimental Setup.}
The neural net used consists of a single hidden layer of 50 units for all data sets except \texttt{protein}, which gets 100. 
The results for all of the baselines except for Variational Dropout (VarOut) are quoted from the results reported by the respective papers who introduced them, while the results on the sparse GP are reported via~\citep{bui2016deep}. For VarOut we rely on our own implementation as there are no official results. 
BEDL, BEDL+Reg, and VarOut all share the same initialization scheme for the mean and variance parameters for each weight following the initialization of~\citep{louizos2017multiplicative}, i.e.\ He-Normal for the means and  $\mathcal{N}(-9, 0.001)$ for the log variances. VarOut gets a Normal prior with a precision of $\alpha=1.0$, and all three get an observation precision of $\beta=100$, to encourage them to learn as much of the predictive uncertainty instead of relying on a fixed hyper-parameter. Note that we keep these values fix and data set independent, different to many of the baselines who set them to data set specific values given cross-validations on separate validation subsets.  

Each model is trained with the Adam optimizer with default parameters for 100 epochs with a learning rate of $10^{-3}$, with varying minibatch sizes depending on the data set size.

\subsection{Classification and Out-of-domain Detection.}
The network for this task follows the common LeNet5 architecture with the following modifications. Instead of max-pooling layers after the two convolutional layers, the convolutional layers themselves use a larger stride to mimic the behavior. And for the more complex CIFAR data set the number of channels in the two convolutional layers is increased from the default 20,50 to 192 each, while the number of hidden units for the fully connected layer is increased from 500 to 1000 for that data set following~\citep{gal2015bayesian}. 

Since there are no OOD results on the BNN baselines we compare against, we rely on our own reimplementations of them, ensuring that they each share the same initialization schemes as in the regression setup. 
For DVI we implement the diagonal version and use a sampling-based approximation on the intractable softmax. Each model gets access to five samples whenever it needs to conduct an MC sampling approximation. All models get trained via the Adam optimizer with the default hyperparameters and a learning rate of $10^{-3}$. 
For EDL we rely on the public implementation the authors~\citep{sensoy2018evidential} provide and use their hyperparameters to learn the model. 
Due to the license status of the not-MNIST data conflicting with the affiliation of the authors, we have to change the setup of earlier work, e.g.~\citep{lakshminarayanan2017simple,louizos2017multiplicative, sensoy2018evidential}, using instead Fashion-MNIST as the closest substitute.

\subsection{Comparison to GP variants.} 
We evaluate the impact of local weight realization on prediction performance by comparing BEDL+Reg to GPs with kernels derived from BNNs with global weight realizations~\citep{alonso2019deep, leep2018deep, neal1995bayesian} on MNIST and CIFAR10 data sets.  
It is technically not possible to perform this evaluation in a fully commensurate setup, as these baselines assume infinitely many neurons per layer and do not have weight-specific degrees of freedom.
Furthermore,~\cite{alonso2019deep} perform neural architecture search and~\cite{leep2018deep} use only part of the CIFAR10 training set reporting that the rest does not fit into the memory of a powerful workstation. We nevertheless view the performance scores reported in these papers as practical upper bounds and provide qualitative comparison. For the choice of neural net depth, we take NNGP~\citep{leep2018deep} as a reference and devise a contrarily thin two-layer convolutional BEDL+Reg network. The results and the architectural details are summarized in Table~\ref{tab:gpcomp}. BEDL and BEDL+Reg can reach lower error rates using significantly less computational resources.

\begin{table}
  \centering
  \adjustbox{max width=\columnwidth}{
    \begin{tabular}{lcc}
      \toprule
      &    \textbf{MNIST} & \textbf{CIFAR10}\\
      \midrule
      NNGP      &           $1.21$  & $44.3$\\   
      Convolutional GP     &    $1.17$ & $35.4$ \\   
      ConvNet GP    &          $1.03$ &      -\\
      Residual CNN GP    &     $0.96$ &      -\\ 
      ResNet GP    &           $0.84$ &      -\\ 
      \midrule
      BEDL (Ours)     &       $0.91$     & $34.20$  \\
      BEDL+Reg (Ours)    &           ${\bf 0.63}$  & ${\bf 32.47}$\\ \bottomrule
    \end{tabular}  
  }
  \caption{\textbf{Comparison to GP Variants.} Test error in~$\%$ on two image classification tasks. BEDL reaches lower error rate than previously proposed neural net based GP constructions by two convolutional layers with $96$ filters of size $5 \times 5$ and stride $2$. BEDL converges in 50 epochs, amounting to circa 30 minutes of training time on a single GPU. The GP alternatives have been reported to have significantly larger time and memory requirements. The GP results are cited from~\citep{alonso2019deep}
  }\label{tab:gpcomp}
\end{table}

\paragraph{Experimental Setup.}
The results for the baselines are taken from the respective original papers. The nets for BEDL and BEDL+Reg consist of two convolutional layers with 96 filters of size $5\times 5$ and a stride of 5. They are trained until convergence (50 epochs) using Adam with the default hyperparameters and a learning rate of $10^{-3}$.

\subsection{Computational cost.} 
Table~\ref{tab:cost} summarizes the computational cost analysis of the considered approaches.  MC Dropout and VarOut can quantify uncertainty only by taking samples across weights, which increases the prediction cost linearly to the sample count. DVI and BEDL+Reg perform the forward 
pass during both training and prediction time via analytical moment matching at double and triple costs, respectively. Both methods have sampling costs for intractable likelihoods.\footnote{Even this sampling step could be avoided by a suitable Taylor approximation, see e.g. Appendix B.4/B.5 in \citep{wu2018fixing}. As the added approximation error was more detrimental to model performance than a cheap MC approach in preliminary experiments, we stay with the latter for both.}
BEDL+Reg may also have another additive per-data-point sampling cost for calculating intractable functional mapping regularizers. Favorably, both of these overheads are only additive to the forward pass cost, i.e. sampling time is independent of the neural net depth, hence they do not set a computational bottleneck. The training and prediction cost of BEDL+Reg is three times EDL which builds on deterministic neural nets. However, it provides substantial improvements in both prediction accuracy and uncertainty quantification.

\begin{table}[h]
  \centering
  \adjustbox{max width=\columnwidth}{
    \begin{tabular}{lll}
      \toprule
      & {\bf Training per iteration} & {\bf Prediction}\\
      \midrule
      MC Dropout  & $\mcO\big((F+L)S\big)$ & $\mcO\big((F+L)S\big)$  \\
      VarOut &   $\mcO\big(2(F+L)S+R (W/N)\big)$  &  $\mcO\big(2(F+L)S\big)$ \\
      DVI  & $\mcO\big(2F+SL+R (W/N)\big)$  & $\mcO\big(2F+SL\big)$\\
      EDL & $\mcO\big(F+L\big)$ & $\mcO\big(F+L\big)$  \\
      BEDL & $\mcO\big(3F+SL\big)$ & $\mcO\big(3F+SL\big)$\\
      BEDL+Reg & $\mcO\big(3F+S(L+R)\big)$ & $\mcO\big(3F+SL\big)$\\
      \bottomrule
  \end{tabular}}
  \caption{\textbf{Computational Cost.} Per data point computational cost analysis in FLOPs. {\bf F:} Forward pass cost of a deterministic net. {\bf W:} Number of weights in the net. {\bf L:} Analytical calculation cost for the exact or approximate likelihood or the loss term. {\bf S:} Number of samples taken. {\bf R:} The cost of the regularization term per unit (weight or data point).  }\label{tab:cost}
\end{table}

\end{document}